\documentclass{IEEEoj-data}
\usepackage{cite}
\usepackage{amsmath,amssymb,amsfonts}
\usepackage{algorithmic}
\usepackage{graphicx,color}
\usepackage{textcomp}
\usepackage{hyperref}
\usepackage{balance}
\usepackage{booktabs}
\usepackage{svg}
\usepackage{subcaption}
\usepackage{graphicx}
\def\BibTeX{{\rm B\kern-.05em{\sc i\kern-.025em b}\kern-.08em
    T\kern-.1667em\lower.7ex\hbox{E}\kern-.125emX}}
\AtBeginDocument{\definecolor{ojcolor}{cmyk}{0.93,0.59,0.15,0.02}}
\definecolor{electricpurple}{rgb}{0.75, 0.0, 1.0}
\usepackage{xcolor} 
\definecolor{emerald}{rgb}{0.31, 0.78, 0.47}

\begin{document}
\title{\textcolor{black}{AI‑Driven Generation of Old English: A Framework for Low‑Resource Languages}}

\author{
  Rodrigo Gabriel Salazar Alva\authorrefmark{1},
  Matías Núñez\authorrefmark{2,1},
  Cristian López\authorrefmark{1},
  Javier Martín Arista\authorrefmark{3}\\[1ex]
  \authorrefmark{1}\textit{Universidad de Ingeniería y Tecnología (UTEC), Lima, Perú}\\
  \authorrefmark{2}\textit{Consejo Nacional de Investigaciones Científicas y Técnicas (CONICET), Buenos Aires, Argentina}\\
  \textit{Instituto de Investigaciones en Biodiversidad y Medioambiente (INIBIOMA), Universidad Nacional del Comahue, Bariloche, Argentina}\\
  \authorrefmark{3}\textit{Universidad de La Rioja, Logroño, España}
}

\begin{abstract}

Preserving ancient languages is essential for understanding humanity’s cultural and linguistic heritage, yet Old English remains critically under-resourced, limiting its accessibility to modern natural language processing (NLP) techniques. We present a scalable framework that uses advanced large language models (LLMs) to generate high-quality Old English texts, addressing this gap. Our approach combines parameter-efficient fine-tuning (Low-Rank Adaptation, LoRA), data augmentation via backtranslation, and a dual-agent pipeline that separates the tasks of content generation (in English) and translation (into Old English). Evaluation with automated metrics (BLEU, METEOR, and CHRF) shows significant improvements over baseline models, with BLEU scores increasing from 26 to over 65 for English-to-Old English translation. Expert human assessment also confirms high grammatical accuracy and stylistic fidelity in the generated texts. Beyond expanding the Old English corpus, our method offers a practical blueprint for revitalizing other endangered languages, effectively uniting AI innovation with the goals of cultural preservation.

\end{abstract}


\maketitle

\section*{Introduction}

Language is among the most profound tools of human civilization, embodying cultural heritage, historical knowledge, and intellectual evolution. Studying ancient languages reveals the origins of societies and their ties to the present. Among these, Old English holds a unique position as the earliest form of the English language, serving as the foundation for contemporary English. Spoken between the 5th and 11th centuries of the Christian Era. it represents a rich tapestry of linguistic complexity, although it remains critically under-resourced in today's paradigm of widespread generative artificial intelligence.

Despite its historical significance, Old English faces issues common to many low-resource languages (LRLs): scarce annotated data, limited accessibility, and an insufficient digital corpus for computational analysis. The written records comprise only three million words—orders of magnitude smaller than corpora available for contemporary languages. This limitation hinders not only linguistic research but also the application of state-of-the-art Natural Language Processing (NLP) techniques, which often depend on large, high-quality datasets \cite{sommerschield2023machine}. 

While powerful Large Language Models (LLMs) have revolutionized NLP, they are not directly capable of generating high-quality Old English texts due to several limitations. First, most models are English-centric, as they have been pre-trained on a predominantly English corpus. Furthermore, even in multilingual models, the lack of sufficient representation of low-resource languages like Old English leads to incomplete and inconsistent understanding. In the case of Old English, its unique linguistic features, such as its complex case system, free-word order, and Germanic vocabulary, further exacerbate the issue, resulting in stylistically or grammatically inaccurate outputs. These challenges call for targeted training and augmentation strategies to adapt LLMs for Old English.

This work aims to bridge the resource gap for Old English by employing large language models (LLMs) to expand its digital corpus through synthetic data generation. Unlike traditional methods that rely on manual curation or scarce text sources, we propose an innovative framework that leverages advanced LLM architectures to generate high-quality Old English texts.

The primary objective of this study is to adapt a pretrained large language model for data generation tasks in Old English through a systematic fine-tuning process that leverages the limited available Old English data and advanced training techniques. This approach enables the model to generate syntactically well-formed and semantically accurate Old English text. Furthermore, to guide the stylistic and contextual quality of the generated output, the proposed framework uses a dual-agent architecture: a generative agent constructs coherent Modern English prompts, and a translation agent renders these into high-quality Old English.

Evaluation was conducted using both automated metrics, such as BLEU, CHRF, and METEOR, and human evaluation by experts in Old English. Texts were rated on grammatical accuracy (inflection and word order), lexical selection (attestedness), and semantic coherence, with only high-quality outputs incorporated into the extended corpus. This refinement ensured the data met both linguistic and computational standards.

To realize this goal, our approach begins by adapting a state-of-the-art language model through a carefully staged training pipeline. The process starts with domain adaptation, where the model is progressively exposed to authentic Old English data—however scarce—alongside carefully selected Modern English examples, leveraging advanced techniques such as domain-adaptive pretraining and efficient fine-tuning. Synthetic data generation is then achieved via a dual-agent system: one agent crafts stylistically consistent Modern English prompts, while a specialized translation agent renders these into fluent Old English, enriched with context and guided by few-shot learning. By iteratively expanding and refining the training data, our methodology enables the model to produce Old English texts with unprecedented fidelity, opening the door to scalable digital preservation and revitalization for other low-resource languages.

The contributions of this work are manifold. It expands the Old English corpus and provides a scalable framework for low-resource languages (LRLs). Moreover, this approach serves as a replicable template for other under-resourced languages, fostering their preservation and study. Beyond linguistics, this work unites computational techniques with cultural heritage preservation, thus contributing to a movement that democratizes access to linguistic resources and ensures the survival of underrepresented languages. By demonstrating how LLMs can effectively address low-resource language challenges, we provide valuable insights for researchers at the juncture of machine learning, humanities, and technology—showing how the same tools shaping the future of artificial intelligence can be repurposed to illuminate the linguistic past of mankind.

\section*{Methods}

This study addresses the challenge of expanding the Old English corpus by adopting a structured, multi-stage methodology. We reconceptualize synthetic data generation as a machine translation task, leveraging the strengths of models trained on resource-rich languages to manage the linguistic complexities inherent to Old English. The workflow comprises three primary stages: data preparation, model training, and synthetic data generation. Throughout these stages, we utilize advanced machine learning  techniques—including Low-Rank Adaptation (LoRA) for efficient fine-tuning, back-translation for robust data augmentation, and a dual-agent architecture for targeted generation and translation. \footnote{Code and resources: \url{https://github.com/tux550/OldEnglish-LLM}}

\subsection*{Data Preparation}

To establish a strong foundation for training, a diverse dataset was curated. The primary source was the Dictionary of Old English Corpus (DOEC), which contains the complete written records of the language, around 3,000 texts comprising 3 million words \cite{healey2004dictionary}. The DOEC has gathered prose and poetry texts, belonging to a wide range of styles: religious texts (homilies and sermons, biblical translations, hagiographies, and liturgical texts), legal texts, historical and chronicle texts, literary texts (poetry, riddles and wisdom literature), scientific and medical texts, glosses and glossaries, letters and administrative texts (charts and wills), proverbs and maxims and miscellaneous texts. Additionally, an annotated subcorpus of the DOEC was employed to provide the model with examples of translations \cite{martin_arista2023parcoroev3}.  The Bosworth-Toller Anglo-Saxon Dictionary \cite{bosworth1972dictionary} served as a complementary dataset, offering definitions and usage contexts for words in Old English.

The collected texts underwent rigorous standardization to ensure consistency across sources. Issues such as non-standard characters, diacritics, and linguistic ambiguities were addressed using a modified version of  \textit{The Classical Language Toolkit} \cite{johnson-etal-2021-classical}, which was adapted to standardize punctuation and character representation. Additionally, low-quality samples were filtered out. The finalized dataset was split into training, validation, and testing subsets, with monolingual Old English texts separately prepared for data augmentation tasks such as back-translation.

\subsection*{Model Training}

\begin{figure}[h]
  \centering
  \fbox{
        \includegraphics[width=0.9\linewidth]{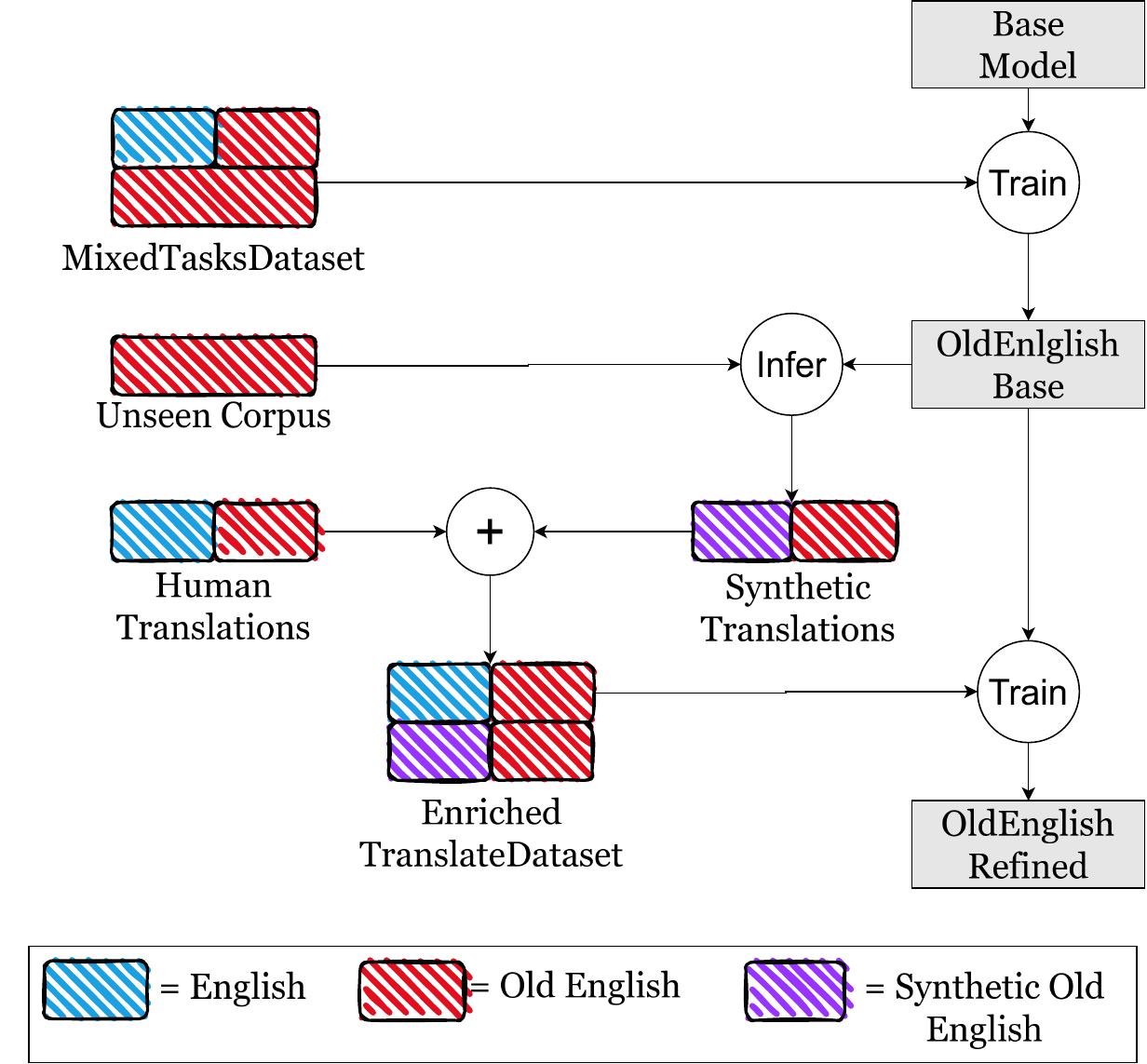}
  }
  \caption{
    \textbf{Schematic of the proposed training pipeline for Old English text generation.} The process begins with a base model trained using \textit{Efficient Task-Similar Domain-Adaptive Continual-Pretraining} on a mixed dataset of English and Old English segments (blue and red, respectively). Synthetic Old English translations (purple) are generated from an unseen English corpus via inference. These are then combined with human-annotated parallel data to create an enriched translation dataset. In a final step, the model is further specialized (fine-tuned) using this combined dataset, resulting in a refined model with improved fluency and linguistic fidelity. Color coding highlights English (\textcolor{blue}{blue}), Old English (\textcolor{red}{red}), and synthetic Old English (\textcolor{electricpurple}{purple}) data at each stage.
  }
  \label{fig:trainflow}
\end{figure}

The training process adapts the language model to Old English through a progressive, multi-stage approach, combining Domain-Adaptive Pretraining (DAPT) and Task-Adaptive Pretraining (TAPT) to achieve greater linguistic and stylistic accuracy~\cite{gururangan2020dontstoppretrainingadapt}. As illustrated in Figure~\ref{fig:trainflow}, this process systematically transitions the model into the new language domain, enhancing both grammatical fidelity and alignment with Old English stylistic norms.

For clarity and consistency, we use the ISO 639-3 codes \texttt{ANG} (Old English) and \texttt{ENG} (Contemporary English) to refer to these languages in all prompt templates, tables, and diagrams throughout this section and the remainder of the paper.

\subsubsection*{Phase 1: Domain Adaptation}

The initial phase of our training pipeline centers on domain adaptation, where the language model is progressively exposed to Old English as its new target domain. This step is crucial for equipping the model with a foundational grasp of Old English structures and vocabulary, ensuring it can both interpret and generate authentic language fragments.

Due to the scarcity of high-quality Old English resources, traditional Domain-Adaptive Pretraining (DAPT)~\cite{gururangan2020dontstoppretrainingadapt} is not feasible at scale. To overcome this limitation, we implement an Efficient Task-Similar Domain-Adaptive Continual-Pretraining strategy~\cite{xie-etal-2024-efficient}. This approach leverages the model’s existing proficiency in Modern English to facilitate a smoother and more data-efficient transition to Old English, thereby maximizing the utility of the available resources.

Specifically, we define four related tasks that guide the model’s adaptation:

\begin{enumerate}
    \item \textbf{Text Completion:} The model is prompted to complete an Old English fragment, reinforcing its ability to generate fluent continuations.
    \item \textbf{Forward Translation:} The model translates Modern English fragments into Old English, directly targeting the core generation task.
    \item \textbf{Back Translation:} The model translates Old English fragments back into Modern English, promoting bidirectional understanding and reducing overfitting.
    \item \textbf{Crossed Definition:} The model provides Modern English definitions for Old English words, strengthening lexical alignment between the languages.
\end{enumerate}

Examples of these prompt formats are provided in Table~\ref{tab:examples_prompts}, illustrating how each task contributes to the overall domain adaptation.

For this phase, we initialize from the Llama-8b checkpoint~\cite{dubey2024llama3herdmodels} and employ Low-Rank Adaptation (LoRA)~\cite{hu2021lora} for computationally efficient training. The resulting checkpoint, adapted for Old English, is referred to as \textit{OldEnglishBase}.

\begin{table*}
\centering
\renewcommand{\arraystretch}{1.5} 
\caption{\textbf{Illustrative examples of tasks designed for Old English adaptation.} This table presents one example per task used in the domain adaptation phase of the proposed training method. These tasks exemplify how knowledge of contemporary languages can be leveraged to facilitate adaptation to low-resource target languages such as Old English.}
\begin{tabular}{@{}cp{7cm}p{7cm}@{}}
\toprule
\textbf{Task}
    & \textbf{Input}
    & \textbf{Output} \\
\midrule
\textbf{Text Completion}
    & \oldenglish{[ANG]and fæste mann þærto swa fela daga swa þærto}
    & \oldenglish{fæstene arærde wæron and þenung togesett is.[/ANG]} \\
\textbf{Forward Translation}  
    & \oldenglish{[INST]Translate the following English fragment to Anglo-Saxon[/INST][EN]xxi. what kind of men the deans of the monastery must be.[/EN]}
    & \oldenglish{[ANG]xxi. hwilce mynstres teoðingealdras beon sceolon.[/ANG]} \\
\textbf{Back Translation}  
    & \oldenglish{[INST]Translate the following Anglo-Saxon fragment to English[/INST][ANG]se oðer him andwirde and cwæð :[/ANG]}
    & \oldenglish{[EN]the second answered him and said :[/EN]} \\  
\textbf{Crossed Definition}
    & \oldenglish{[INST]What is the English definition of the following word in Anglo-Saxon?[/INST][ANG]getoge[/ANG]}
    & \oldenglish{[EN]A tugging; contractio; contraction; convulsio; convulsion; cramp; spasm; spasmus[/EN]} \\
\bottomrule
\end{tabular}
\label{tab:examples_prompts}
\end{table*}

\subsubsection*{Phase 2: Task Specialization}

Building on the \textit{OldEnglishBase} model obtained from the domain adaptation phase, the second stage of training focuses on task specialization. In this step, we introduce the technique of backtranslation~\cite{edunov2018understandingbacktranslationscale}, a widely adopted data augmentation method in low-resource neural machine translation.

Backtranslation works by taking monolingual Old English text from the corpus and translating it into Modern English using the model’s strong reverse translation capability. Because the base model is English-centric, it reliably produces high-quality Modern English translations from Old English input at this stage~\cite{LDPxuan2023}. These newly generated Modern English sentences are then paired with their original Old English counterparts to form synthetic parallel corpora. This expanded dataset introduces greater lexical and stylistic variety, which is critical for further fine-tuning the model’s ability to generate Old English from Modern English prompts.

During the synthetic data generation process, we feed the model previously unseen fragments from the monolingual Old English corpus and use greedy inference~\cite{edunov2018understandingbacktranslationscale} to produce accurate Modern English translations. By combining these synthetic parallel pairs with the existing human-annotated examples, we assemble a substantially richer and more diverse training set for forward translation.

Fine-tuning the model on this enriched dataset yields the \textit{OldEnglishRefined} model. This final checkpoint demonstrates markedly improved capabilities, generating Old English texts that are both more coherent and more contextually faithful to the original Modern English input.

\subsection*{Synthetic Data Generation}

Following the domain adaptation and task specialization phases—which collectively produce a robust model for Old English translation—the final stage addresses the persistent scarcity of Old English corpora by generating additional, high-quality synthetic data. To maximize both linguistic diversity and fidelity, we implement a dual-agent architecture that divides the generation process into two coordinated steps: fragment mutation and translation.

\begin{figure}[h]
  \centering
  \fbox{
    \includegraphics[width=0.9\linewidth]{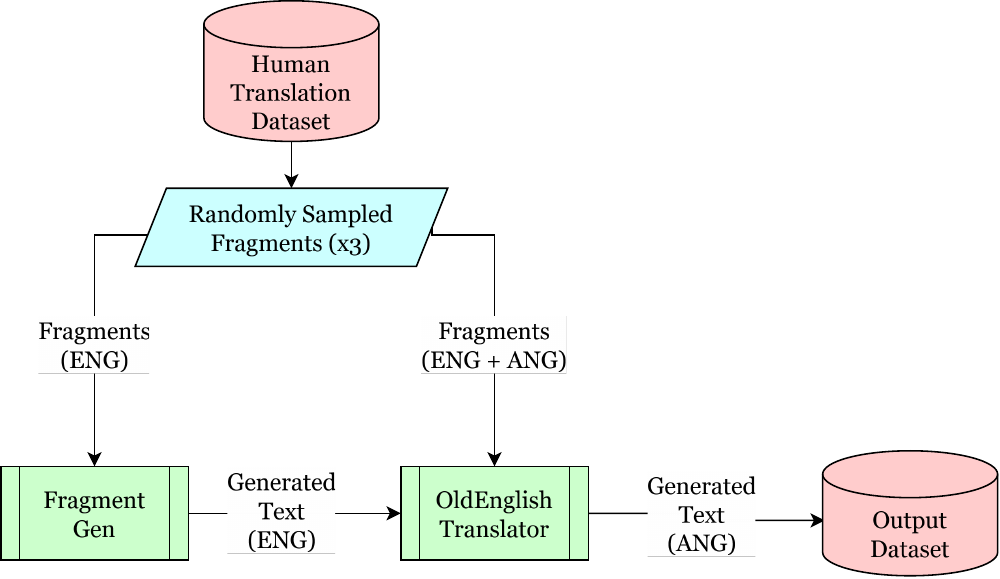}
  }
  \caption{
    \textbf{Dual-agent synthetic data generation pipeline.} Datasets (\textcolor{red}{red}) provide randomly sampled fragments (\textcolor{blue}{blue process}) to two collaborating agents (\textcolor{emerald}{green}). First, \textit{FragmentGen} creates new Modern English text using few-shot prompts for stylistic alignment. Then, the \textit{OldEnglishTranslator} converts these into Old English, leveraging bilingual context. The result is a high-quality output dataset of synthetic Old English text. This architecture maximizes linguistic diversity and coherence by combining guided mutation and translation in a controlled, multi-agent workflow.
  }
  \label{fig:agents}
\end{figure}

In this pipeline, shown schematically in Figure~\ref{fig:agents}, the agents interact with both authentic and synthetic datasets, with each agent performing a distinct but complementary role. Randomly sampled fragments from the reference corpus are provided as contextual anchors for the agents, ensuring that the generated sentences remain stylistically consistent and lexically relevant to real Old English usage.

\subsubsection*{Fragment Generator Agent}

The first agent, \textit{FragmentGen}, is responsible for producing new text fragments in Modern English. To guide this process, the agent receives a selection of example sentences randomly drawn from the Dictionary of Old English Corpus (DOEC), which serves as a stylistic template and vocabulary guide. This context-driven prompting strategy reduces the likelihood of out-of-vocabulary or domain-inappropriate content, thereby facilitating more effective downstream translation. Since this agent operates exclusively in Modern English, it can leverage advanced general-purpose language models without requiring any Old English expertise. For this study, we employ GPT-4o-mini for its strong generative capabilities and controllability.

\subsubsection*{Translation Agent}

The second agent, \textit{OldEnglishTranslator}, utilizes the \textit{OldEnglishRefined} model to translate the Modern English fragments into Old English. To preserve grammatical accuracy and stylistic coherence, the agent applies few-shot prompting, drawing on bilingual pairs originally sampled by the \textit{FragmentGen} agent. Greedy decoding is used for inference, ensuring reproducibility and minimizing unnecessary variation in the outputs.

By integrating these two specialized agents into a unified pipeline, we can efficiently generate synthetic Old English fragments that are not only linguistically diverse but also adhere closely to authentic stylistic and lexical norms. This process substantially augments the available training data and further improves model performance for both translation and text generation tasks in low-resource settings.

\subsection*{Evaluation}
The evaluation process combined automated metrics and human assessments to ensure reliability and quality.

For the automated metrics we employed BLEU \cite{papineni2002bleu}, METEOR \cite{banerjee2005meteor} and CHRF \cite{popovic2015chrf}. These metrics function as quantitative proxies for human judgment, comparing the generated translation to reference human translations. BLEU evaluates precision at the n-gram level, assessing how much of the generated text overlaps with the reference. METEOR improves on this by incorporating recall and using techniques such as stemming and synonym matching for better alignment with human evaluation. CHRF, in contrast, evaluates translations based on character-level precision and recall.

Regarding human evaluation, expert reviewers rated the outputs on three criteria: grammatical accuracy (inflection and word order), stylistic fluency (lexical choice), and contextual coherence. The evaluation was conducted on a 10-point scale, with a threshold score of 7 set to determine whether a text is appropriate for incorporation into the extended corpus.

\section*{Results and Discussion}

This study provides compelling evidence that large language models (LLMs) can effectively address the challenge of corpus expansion for low-resource languages, with Old English serving as a representative case study. Through a systematic combination of advanced model fine-tuning, targeted data augmentation, and a modular, task-specific pipeline, we achieved significant improvements in both translation quality and linguistic fidelity. Our results showcase not only substantial performance gains over baseline models but also illuminate the specific strengths and persistent limitations encountered when adapting LLMs to a historically and linguistically complex low-resource setting.

\subsection*{Model Fine-Tuning and Performance}

\begin{table}[h]
    \centering
    \caption{Evaluation of baseline models in translation tasks using BLEU, CHRF, and METEOR scores. All displayed values are reported on a 0-100 scale in with higher scores indicate closer correspondence with reference human translations.}
    \begin{tabular}{@{}l|ccc@{}}
        \toprule
        \textbf{Modelo} & \textbf{BLEU} & \textbf{CHRF} & \textbf{METEOR} \\ \midrule
        \multicolumn{4}{c}{$EN \rightarrow ANG$} \\ 
        \midrule
        Llama3.1 8B & \textbf{29.13} & \textbf{25.51} & \textbf{18.95} \\
        Mistral 7B  & 24.51 & 22.61 & 16.12 \\ 
        \midrule
        \multicolumn{4}{c}{$ANG \rightarrow EN$} \\ 
        \midrule
        Llama3.1 8B & \textbf{46.87} & \textbf{38.92} & \textbf{44.20} \\
        Mistral 7B  & 43.94 & 36.31 & 40.27 \\ 
        \bottomrule
    \end{tabular}
    \label{tab:eval_base_models}
\end{table}

\begin{table*}[ht]
\centering
\renewcommand{\arraystretch}{1.5}
\caption{\textbf{Common translation errors identified in the base models during forward translation evaluation.} The examples illustrate issues such as looped generation, non-translated segments, and hallucinated vocabulary.}
\begin{tabular}{@{}cp{7cm}p{7cm}@{}}
\toprule
\textbf{Error type}
    & \textbf{Input}
    & \textbf{Generation} \\
\midrule
\textbf{Looped generation}  
    & That is, at stoke twelve hides; and they have also given me thirty mancuses of gold, and that i free the monastery from all the taxes which belong to the lord of the people (...)
    & Thæt is, at stoc twelve hides; and hi hafa meth meth hund mancusa golds, and thæt ic monæstære fræolice fræolice fræolice fræolice fræolice fræolice fræolice fræolice (...) \\  
\textbf{Non-translated segments}
    & the roman people first called it by that name, because on the first day of the month he established the roman empire and overthrew those who formerly had destroyed it (...)
    & the roman people first called it by that name, because on the first day of the month he established the roman empire and overthrew those who formerly had destroyed it (...)\\
\textbf{Vocabulary hallucination}
    & first he killed amyntas, his aunt's son, and afterwards his brother, and then pannenio, his general, and then philotas (...)
    & he gefeng amyntas, his modorbrothor, and eft pannenio, his herefore, and eft philotas (...) \\

\bottomrule
\end{tabular}
\label{tab:common_issues}
\end{table*}

The initial evaluation of baseline models—LLaMA-8b~\cite{dubey2024llama3herdmodels} and Mistral-7B~\cite{jiang2023mistral7b}—revealed critical limitations in their ability to process Old English. Both models achieved moderate results in reverse translation tasks (Old English to Contemporary English), yet their performance in direct translation (Contemporary English to Old English) lagged significantly behind (Table~\ref{tab:eval_base_models}). This pronounced asymmetry highlights the inherent difficulties of generating accurate and fluent text in a low-resource language characterized by complex grammatical structures.

Among the baselines, LLaMA-8b consistently outperformed Mistral-7B across all evaluation metrics—BLEU, CHRF, and METEOR—showing an average improvement of 2 to 5 points (Table~\ref{tab:eval_base_models}). Nonetheless, both models commonly exhibited translation errors, including repetitive or looped text, untranslated segments, and stylistic inconsistencies (Table~\ref{tab:common_issues}). These recurring issues underscored the necessity of further fine-tuning and methodological innovation to effectively adapt LLMs for high-fidelity Old English text generation.

\begin{table}[h]
    \centering
    \caption{
      \textbf{Translation performance of the \textit{OldEnglishBase} model at different training epochs, evaluated using BLEU, CHRF, and METEOR metrics (all scaled 0–100).} Higher values indicate greater similarity to reference human translations. Results are reported for both Modern English to Old English (\texttt{EN}~$\rightarrow$~\texttt{ANG}) and Old English to Modern English (\texttt{ANG}~$\rightarrow$~\texttt{EN}) directions. Performance improves up to 3 epochs, after which overfitting and catastrophic forgetting reduce scores, especially for reverse translation.
    }
    \begin{tabular}{@{}l|ccc@{}}
        \toprule
        \textbf{Epochs} & \textbf{BLEU} & \textbf{CHRF} & \textbf{METEOR} \\ \midrule
        \multicolumn{4}{c}{$EN \rightarrow ANG$} \\
        \midrule
        1           & 59.39         & 50.83         & 50.13          \\
        3           & \textbf{60.73} & \textbf{55.68} & \textbf{54.19}  \\
        5           & 55.43         & 52.49         & 53.76          \\ 
        \midrule
        \multicolumn{4}{c}{$ANG \rightarrow EN$} \\ 
        \midrule
        1           & \textbf{66.54} & \textbf{60.24} & \textbf{65.91}  \\
        3           & 59.27         & 56.98         & 58.80          \\
        5           & 52.30         & 52.30         & 52.48          \\ 
        \bottomrule
    \end{tabular}
    \label{tab:eval_boot_epoch}
\end{table}

The refinement process demonstrated measurable improvements.

\textit{OldEnglishBase} served as a robust initial model capable of capturing Old English vocabulary and grammar and demostrated a significant improvement over the baseline model in both direct and reverse translation tasks, with score increases of +35 points and +20 points respectively, effectively doubling them (Table \ref{tab:eval_boot_epoch}). However, the improvements quickly stagnated after just one epoch of training, with diminishing returns in direct translation and signs of catastrophic forgetting of its Contemporary English domain for inverse translation.

\begin{table}[h]
    \centering
    \caption{
      Translation performance comparison for different trained models on Modern English~$\rightarrow$~Old English (\texttt{EN}~$\rightarrow$~\texttt{ANG}) and Old English~$\rightarrow$~Modern English (\texttt{ANG}~$\rightarrow$~\texttt{EN}) tasks. Metrics (BLEU, CHRF, and METEOR) are scaled 0–100; higher scores indicate greater similarity to human reference translations. Results show substantial gains moving from the baseline Llama model to \textit{OldEnglishBase}, with further improvements achieved by the \textit{OldEnglishRefined} model, especially in direct translation to Old English.
    }
    \begin{tabular}{@{}l|ccc@{}}
        \toprule
        \textbf{Model} & \textbf{BLEU} & \textbf{CHRF} & \textbf{METEOR} \\ \midrule
        \multicolumn{4}{c}{$EN \rightarrow ANG$} \\ 
        \midrule
        Llama           & 25.94         & 22.97         & 17.43          \\
        \textit{OldEnglishBase} & 59.99         & 51.65         & 51.01          \\
        \textit{OldEnglishRefined}  & \textbf{65.41} & \textbf{57.82} & 57.40          \\
        \midrule
        \multicolumn{4}{c}{$ANG \rightarrow EN$} \\ 
        \midrule
        Llama            & 47.68         & 39.55         & 44.03          \\
        \textit{OldEnglishBase} & 69.41         & 63.20         & 67.45          \\
        \textit{OldEnglishRefined}  & \textbf{69.89} & \textbf{63.54} & \textbf{68.50} \\
        \bottomrule
    \end{tabular}
    \label{tab:synth_evaluacion_eval}
\end{table}

\begin{figure*}[h]
    \centering
    \fbox{
      \begin{minipage}{0.9\textwidth}
        \begin{subfigure}[b]{0.48\textwidth}
            \centering
            \includegraphics[width=\linewidth]{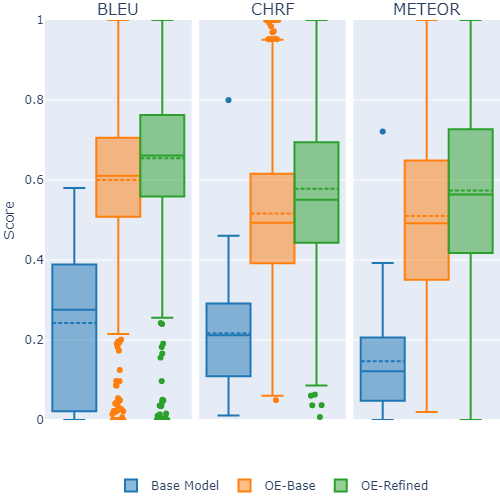}
            \caption{Forward translation ($EN \rightarrow ANG$)}
            \label{fig:synth_ft}
        \end{subfigure}
        \hspace{0.03\textwidth}
        \begin{subfigure}[b]{0.48\textwidth}
            \centering
            \includegraphics[width=\linewidth]{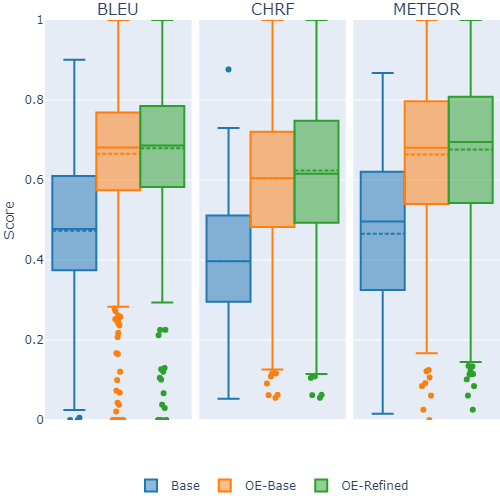}
            \caption{Back translation ($ANG \rightarrow EN$)}
            \label{fig:synth_bt}
        \end{subfigure}
      \end{minipage}
    }
    \caption{
        \textbf{Translation performance distributions for base and trained models.}
        Box plots show the distributions of BLEU, CHRF, and METEOR scores for three model stages (Base, OE-Base, and OE-Refined) on both translation directions: (a) Modern English $\rightarrow$ Old English and (b) Old English $\rightarrow$ Modern English. Scores are normalized to [0, 1]. The dotted line within each box indicates the mean.
        \textbf{Analysis:} For both directions, there is a clear performance progression: OE-Base models (\textcolor{orange}{orange}) show marked gains over the Base (\textcolor{blue}{blue}), with further improvements from OE-Refined (\textcolor{emerald}{green}). This is reflected in higher medians, means, and upper quartiles across all metrics. Notably, the score distribution for OE-Refined is more compact, indicating more consistent high-quality outputs. Forward translation benefits most, with especially large gains in BLEU and METEOR. Outliers are reduced as training progresses, underscoring improved robustness.
    }
    \label{fig:synth_t}
\end{figure*}

The refinement of this model through the technique of backtranslation proved to be key in overcoming the stagnation. \textit{OldEnglishRefined} excelled in producing fluent and contextually appropriate Old English text, showing improvements of approximately 6 points across all metrics for direct translation.  It is important to note that this process specializes the model for direct translation, which is why the gains in reverse translation are minimal.

\begin{table*}[ht]
\centering
\renewcommand{\arraystretch}{1.5}
\caption{Improvement in zero-shot translation quality. This table illustrates the progressive enhancement of translation quality in a zero-shot setting, highlighting the evolution of outputs as the model refines its understanding of the target language.}
\begin{tabular}{@{}ccp{14cm}@{}}
\toprule
\textbf{Source}
    & \textbf{BLEU}
    & \textbf{Text} \\
\midrule
\textbf{English} & - & that is, at stoke twelve hides ; and they have also given me thirty mancuses of gold, and that i free the monastery from all the taxes which belong to the lord of the people, little or great, known or unknown, except.the'angild'to others, aud fortress-work, and 'fyrdsoon,'and bridge-work ...  \\
\textbf{Human} & - & \oldenglish{ðæt is æt stoce twelf hida, and ðritig mancusa goldes hio sealdan eac me, and ic and mynster fram æghwelcum gafolum gefreoge ðe to ðiode hlafarde belimpeð, littles oððe micles, cuðes ge uncuðes, butan angilde wiðoðrum, and fæstengewerce, and fyrdsocne, and brycggeweorce ... }
  \\
\textbf{Llama 7b} & 45 & \oldenglish{beoð þa twelf hida landes, and hæfiað me þa þritig mancusas goldes, and þa þe ic freoðe þone minan monasterium of allum þam scotum þe to þam hlafordes þe þa folc beoð, litel oððe micel, þe þe ic wite oððe þe ic ne wite, butan þam'angild'to oþrum, and 'fyrdson', and 'bricg-weroc'.}  \\
\textbf{OldEnglishBase} & 62 &  \oldenglish{ðæt is æt stoce twelf hida, and heom hæbbe ic eac geunnen xxx mancus goldes, and ic freo ðæt mynstre of eallum ðam gafolum ðe to ðam folcgere belimpað, lytle oððe mycele, cuðe oððe uncuðe, butan angilde to oðrum, and fæstenweorce, and fyrdsocne, and brycggeworce.}  \\
\textbf{OldEnglishRefined} & 65 & \oldenglish{ðæt is æt stoce twelf hida, and hæbbe ic eac geunnen ðæt man freoge ðæt mynstre of eallum ðam gafoles ðe to ðam folces hlaforde belimpað, lytles oððe micles, cuðes oððe uncuðes, butan angilde wið oðrum, and fæstenweorce, and fyrdsocne, and brycggeweorce.}
  \\

\bottomrule
\end{tabular}
\label{tab:translation_evolution}
\end{table*}

The gains obtained by our model align with previous applications of backtranslation in large language models such as the \textit{Linguistically Diverse Prompting} \cite{LDPxuan2023} approach, which reported increases of up to 8 points in the chrF++ metric. Notably, our model begins with a baseline over twice as high due to the domain adaptation step, making substantial improvements more difficult to achieve. Despite this, we attained gains comparable to those of prior work, highlighting the effectiveness of this approach.

\subsection*{Synthetic Data Generation}

The dual-agent pipeline, comprising  \textit{FragmentGen} and  \textit{OldEnglishTranslator}, played a pivotal role in addressing the limited availability of annotated Old English data.  \textit{FragmentGen} effectively generated coherent and thematically diverse texts in Contemporary English, providing high-quality inputs for the translation agent.  \textit{OldEnglishTranslator}, built on the trained  \textit{OldEnglishRefined} model, translated these texts into Old English with a high degree of grammatical and stylistic fidelity.

\subsubsection*{Expert Linguistic Evaluation}

Expert linguists evaluated the generated texts using four criteria: inflection (including nominal and adjectival declension, verbal conjugation, and subject-verb agreement), word order, lexical choice (attestedness in the surviving texts of Old English), and semantic coherence. Each was rated on a 1–10 scale.

For instance, the segment \oldenglish{"spyr on ðam pocce on eagum; heo sceal niman wad and gearwan, forðam hi ealle andswaredon"} (‘Inquire about the pock in the eyes; she shall take woad and yarrow, for they all answered’) was highly rated for its accurate use of imperative and future verbal forms (\oldenglish{spyr, sceal niman}), demonstrating strong command of Old English morphology. The passage also presents a clear instructional structure and employs medical and technical herb terminology with precision, closely reflecting authentic genre conventions. These strengths are reflected in the scores for inflection (9/10), word order (8/10), and lexical choice (10/10). 

However, the segment received a slightly lower score for semantic coherence (7/10), due to an inadequate continuation of the topic: the causal relationship is unclear, as the last clause (\oldenglish{forðam hi ealle andswaredon}) is incorrectly linked to the main predication. Despite this minor issue, the overall evaluation for this passage was very positive, with an average score of 8.5/10, illustrating the model's ability to generate grammatically robust and lexically rich Old English text.

From a qualitative perspective, manual assessment of the synthetically generated texts underscores the model’s considerable strengths in grammatical structure, particularly in inflection and word order. The system reliably produces complex constructions—such as relative and conditional clauses, and other intricate syntactic patterns—with high fidelity. Furthermore, genre decorum is consistently maintained, as the generated texts adapt stylistic conventions appropriate to different genres. Specialised terminology in religious, legal, and historical contexts is both accurate and contextually appropriate, demonstrating the model’s flexibility across diverse textual traditions.

Despite these notable strengths, semantic coherence emerged as the most frequent challenge in the generated texts. In some cases, anachronistic concepts, such as modern dates or technologies, are retrofitted into Old English, leading to slight incongruities. For example, in the segment \oldenglish{"gif ðu bist getogen on andwerdre tide and on tocyme, ðonne scealt ðu gymene habban to ðære lare on armenia"} (‘if you are trained on present and future tense data, you must pay attention to the teaching methods in Armenia’), references to data and training introduce modern elements that are historically out of place.

Complex narratives can occasionally be oversimplified, and unrelated domains may be inadvertently blended. For instance, in the passage \oldenglish{"wið swile gate tyrdlu on scearpum ecede gesoden and on selfe wisan on gedon"} (‘to overcome swelling, apply sharp vinegar sodden with goats treadles and keep going on good advise’), the second coordinate clause diverges from the main narrative, reducing overall coherence.

Across genres, biblical quotations and religious instruction are typically more convincing and stylistically consistent than medical remedies, land charters, or detailed historical chronicles. While inflection and word order remain strong, the model tends to avoid structurally marked configurations such as non-nominative subjects, non-accusative objects, or pragmatic reorderings (like the V2 rule). Another recurring issue is the loss of co-reference tracking in complex syntactic structures. For example, in the segment \oldenglish{"Ða axodon hi hie hu fela sealfe hæbbe ge. Hi cwædon seofon hræfnes \& an rahdeores"} (‘Then they asked her how many salves have you? She said seven crabs and a roebuck’), the reference point shifts from the singular second person to the plural third person, creating ambiguity.

\begin{table}[h]
    \centering
    \caption{
        Results of expert evaluation for model-generated Old English texts. Each criterion was rated on a 0--10 scale, with higher scores indicating closer alignment with grammatical and stylistic expectations found in authentic Old English sources. The model achieved very high marks for inflection, word order, and lexical choice, while semantic coherence, though strong, remains the most challenging aspect. The overall average reflects robust linguistic performance.The model excels in grammatical accuracy and lexical selection, while semantic coherence remains the main area for improvement.
    }
    \vspace{2mm}
    \begin{tabular}{l c}
        \toprule
        \textbf{Evaluation Criterion}  & \textbf{Average Score} \\
        \midrule
        Inflection (morphology)        & 9.0 \\
        Word Order (syntax)            & 9.0 \\
        Lexical Choice (attestedness)  & 9.1 \\
        Semantic Coherence             & 7.8 \\
        \midrule
        \textbf{Overall Average}       & \textbf{8.7} \\
        \bottomrule
    \end{tabular}
    \vspace{1mm}
    
    \raggedright
    \small
    \label{tab:manual_eval}
\end{table}

 As summarized in Table~\ref{tab:manual_eval}, expert evaluation demonstrates that the model excels in producing Old English with high grammatical accuracy and appropriate lexical choices, as reflected by average scores above 9.0 for inflection, word order, and lexical selection. These results indicate a robust command of Old English morphology, syntax, and vocabulary in the generated texts. However, a slightly lower score for semantic coherence (7.8) highlights a recurring limitation: while the output is structurally and lexically convincing, the model occasionally struggles to maintain deep narrative consistency or capture subtle contextual relationships. This assessment is based on detailed, criterion-specific scoring by specialist linguists familiar with Old English, ensuring that both surface-level correctness and deeper semantic fidelity are carefully measured. Overall, consistently high scores across linguistic categories confirm the effectiveness of our approach in generating plausible Old English, while also revealing areas for future improvement in semantic integration.

\subsection*{Challenges and Limitations}

Despite the notable achievements of this study, several challenges remain. The scarcity of high-quality Old English resources continues to constrain the diversity and complexity of training data. Although data augmentation strategies partially alleviate this limitation, certain linguistic features of Old English—such as its intricate case system and free word order—persistently challenge both model training and text generation.

The use of synthetic data, while effective in expanding the corpus, raises questions about the authenticity and historical fidelity of generated texts. While automated metrics indicate high quality, further validation by expert reviewers and systematic comparison with extant Old English corpora are essential to ensure that outputs are linguistically and historically accurate.

A promising avenue for addressing these challenges is the integration of Retrieval-Augmented Generation (RAG) techniques~\cite{lewis2020rag,lyu2023dataimportance}. By embedding curated fragments from authentic Old English sources into a dedicated vector database, the model can dynamically retrieve contextually relevant passages during both training and synthetic data generation. This retrieval-based grounding not only helps mitigate issues such as hallucinations, repetitive text, and stylistic drift, but also provides real historical references to guide the model’s output. Importantly, RAG can significantly improve semantic coherence and context sensitivity, making it especially valuable for downstream applications such as question answering, where generating accurate and contextually anchored responses is critical.

However, implementing RAG in a low-resource setting introduces additional technical complexities. Determining optimal chunk sizes, establishing effective metadata tagging, and managing the computational overhead of retrieval operations are all crucial to prevent performance bottlenecks and maintain system scalability.

Another pathway to improved model performance lies in the use of more powerful LLMs, such as GPT-4, larger variants of LLaMA, or the DeepSeek R1 model. With greater parameter counts and broader training corpora, these models could offer enhanced contextual understanding and stylistic versatility. Nevertheless, such advancements require significantly more computational resources and risk overfitting to noise in limited data environments. Thus, balancing computational efficiency with model accuracy and historical fidelity remains a central concern for future research.

\subsection*{Broader Implications}

This research demonstrates that advanced NLP and machine learning  techniques, when carefully adapted, can bridge the gap between limited linguistic resources and modern computational methods. By generating and analyzing texts in Old English—a historically significant but underrepresented language—our framework provides valuable tools for linguistic research, cultural preservation, and educational initiatives. The ability to produce high-quality synthetic data opens up new avenues for investigating language evolution, enriching digital archives, and creating interactive tools for scholars and the general public.

Importantly, the dual-agent pipeline and data augmentation strategies presented here are broadly applicable to other endangered or low-resource languages. The trained models not only excel at unsupervised text generation, but also show emergent capacity for high-quality translation and downstream NLP applications, such as Named Entity Recognition (NER) and Part-of-Speech (POS) tagging, especially when fine-tuned or adapted through task-specific prompts. Additionally, the flexible design allows for adaptation to domain-specific text generation or smart data augmentation tasks, supporting a wide range of digital humanities and computational linguistics projects.

Beyond the case of Old English, these methodological innovations—combining efficient fine-tuning, iterative backtranslation, and modular architectures—offer a scalable, cost-effective blueprint for the revitalization and study of other languages facing resource scarcity. Thus, this research not only contributes to linguistic and cultural heritage, but also to the broader democratization of artificial intelligence for societal benefit.

\subsection*{Conclusions}

This work introduces a novel and effective framework for the generation of Old English texts using state-of-the-art transformer-based language models. Through a combination of low-rank adaptation, backtranslation, and a dual-agent pipeline, we significantly expand the available corpus of Old English with linguistically accurate and stylistically plausible texts. Our multi-stage approach—encompassing data preparation, domain adaptation, task specialization, and synthetic data generation—proves capable of capturing the grammatical and stylistic intricacies of the language.

Evaluation results confirm the success of our methodology: the generated texts achieve high scores from both automated metrics (BLEU, METEOR, CHRF) and expert evaluators, especially in grammatical accuracy and lexical choice. However, semantic coherence and context remain areas for ongoing improvement, with challenges such as anachronisms and narrative oversimplification still present in some outputs.

The strategies developed here provide a scalable and replicable path for supporting and revitalizing other low-resource languages. Looking forward, integrating retrieval-augmented generation and leveraging more advanced language models offer promising directions to further enhance both linguistic fidelity and contextual accuracy, ensuring these methods remain at the forefront of computational approaches to cultural preservation.

\section*{Acknowledgements}
We thank the support by 
PID2023-149762NB-100,funded by MCIN /
AEI / 10.13039/501100011033

\bibliographystyle{IEEEtran}
\bibliography{referencias}

\end{document}